\documentclass{article}

\title{Feedforward Initialization for Fast Inference of Deep Generative Networks is Biologically Plausible}
\author{
  Yoshua Bengio$^{1*}$, Benjamin Scellier$^1$, Olexa Bilaniuk$^1$, Jo\~{a}o Sacramento$^\dagger$ and Walter Senn$^\dagger$\\
  $^1$Universit\'{e} de Montr\'{e}al, Montreal Institute for Learning Algorithms\\
  $^\dagger$University of Bern, $^*$ Canadian Institute for Advanced Research\\
}

\newif\iffinal
\finaltrue
\iffinal
\usepackage{vmargin}
\setmarginsrb{3cm}{3cm}{3cm}{3cm}{0cm}{0.8cm}{0cm}{1cm}
\else
\usepackage{nips_2016}
\fi
\usepackage[margin=3cm]{caption}
\usepackage{makeidx} 
\usepackage{graphicx}
\usepackage{algorithm}
\usepackage{algorithmic}
\usepackage{natbib}
\usepackage{caption}
\graphicspath{{.},{figures/}}

\usepackage[T1]{fontenc}      
\usepackage[utf8]{inputenc} 
\usepackage{amsmath}
\usepackage{times}
\usepackage{bm}

\makeindex

\newcommand   \vv{{\bm v}}
\newcommand   \vx{{\bm x}}
\newcommand   \vh{{\bm h}}
\newcommand   \vs{{\bm s}}
\newcommand   \vd{{\bm d}}

\begin{document}

\maketitle


\begin{abstract}
  We consider deep multi-layered generative models such as Boltzmann machines or Hopfield nets in which computation (which implements inference) is both recurrent and stochastic, but where the recurrence is not to model sequential structure, only to perform computation. We find conditions under which a simple feedforward computation is a very good initialization for inference, after the input units are clamped to observed values. It means that after the feedforward initialization, the recurrent network is very close to a fixed point of the network dynamics, where the energy gradient is 0. The main condition is that consecutive layers form a good auto-encoder, or more generally that different groups of inputs into the unit (in particular, bottom-up inputs on one hand, top-down inputs on the other hand) are consistent with each other, producing the same contribution into the total weighted sum of inputs. In biological terms, this would correspond to having each dendritic branch correctly predicting the aggregate input from all the dendritic branches, i.e., the soma potential. This is consistent with the prediction that the synaptic weights into dendritic branches such as those of the apical and basal dendrites of pyramidal cells are trained to minimize the prediction error made by the dendritic branch when the target is the somatic activity. Whereas previous work has shown how to achieve fast negative phase inference (when the model is unclamped) in a predictive recurrent model, this contribution helps to achieve fast positive phase inference (when the target output is clamped) in such recurrent neural models.
\end{abstract}

\section{Introduction}

We are still far from having a theory of how brains learn complex functions
that is both biologically plausible and makes sense from a machine learning
point of view. Unlike many artificial neural networks, the cortex and other
areas of the brain have both feedforward and feedback connections: when
area A sends signals to area B, there usually are also connections from B
to A. Many proposed learning algorithms for such recurrently connected
but non-temporal networks, such as variants of the Boltzmann machine~\citep{Hinton84} or the
contrastive Hebbian algorithm for Hopfield
networks~\citep{Movelland+McClelland91,Xie+Seung-2003} involve two kinds of operations or
``phases'': a ``positive'' phase where the observations are clamped on
visible units and a ``negative'' phase the network is free-running.
To emulate supervised learning (with input units and output units),
the setting would have the inputs clamped in both phases and the outputs
clamped only in the positive phase. The training objective is always
to match behavior in both phases. What makes many of these algorithms
impractical from both a machine learning and biological point of view
is the need for lengthy iterative relaxation to reach either a fixed point
(for deterministic networks) or a stationary distribution (for stochatic
networks), and that applies to both phases and to both the purely
unsupervised and the supervised settings. Note that the networks we
are talking about are recurrent but they are applied to a static
input: this non-temporal recurrence is only used to iteratively
compute a good solution, to perform
inference or to sample from the model.
For example, one runs a Monte-Carlo Markov
chain (MCMC) in the case of Boltzmann machines, or iterates to a fixed point
by going down the model's energy, in
the case of Hopfield networks.
This is biologically implausible because a biological
agent needs to be able to react quickly to a new stimulus. It is also impractical
from a machine learning point of view, because running an MCMC or
iterating to a fixed-point in the inner loop of training could slow
down training considerably. On the other hand, there is plenty of evidence
that very complex functions can be learned by using only back-propagation
to compute gradients, at least in the supervised case,
which requires a single feedforward pass and a single
backward pass. Bringing that kind of speed to non-temporal recurrent networks would
be useful for two reasons:
\begin{enumerate}
  \item These non-temporal recurrent networks can perform
    computations that feedforward networks cannot perform, such as
    filling in for missing values (or sampling from the conditional
    distribution of any subset of variables given any other subset)
    and representing full joint distributions rather than just point predictions.
  \item They better match a basic feature of brains, i.e., the recurrence
    due to feedback connections.
\end{enumerate}

The main contribution of this paper is to show that under the right
conditions, which are not difficult to achieve, a multi-layer recurrent
network can perform inference almost as fast as a feedforward network.
This is achieved by making sure that different dendritic branches
of the same neuron form a mutual prediction, i.e., that the signal coming
from dendritic branch matches, or that they predict
the average output of the various dendritic branches. We show that equivalently, this means that
each pair of successive layers forms a good auto-encoder.
This is interesting because it draws links between feedforward
networks and non-temporal recurrent networks, and because it could provide a
general tool for faster approximate inference.

The first such link between feedforward computation and a non-temporal recurrent network
was probably that made by~\citet{Pineda87,Almeida87}, who show one way to compute gradients
  through the fixed point computation, the equivalent of back-propagation
  for such networks. Unfortunately the resulting algorithm is highly implausible in terms of biology,
  because the neurons would have to perform linear computation in the backprop relaxation phase.
  A related algorithm was proposed by~\cite{Xie+Seung-2003}, requiring the feedback connections
  to have very small weights (in order to obtain a form of linearization of the recurrent computation),
  but showing an exponentially fast decay of the gradient being back-propagated.
  Yet another approach was introduced by~\citet{Scellier+Bengio-arxiv2016}, who propose
  to clamp the target output only a very small distance away from the predicted output:
  in this way the neurons always compute according to the same equations, with no
  need for a different behavior when backpropagating.

\section{Recurrent Stochastic Networks}

Here we review basic notions of recurrent stochastic networks of the kind
for which the proposed approximate inference is applicable, i.e., not meant
to read or generate sequences but rather to reconcile many pieces of information
coming from different parts of the observed input, as well as to reconcile
the observed evidence with the implicit prior held by the model. This paper
is only about the opportunity to perform fast approximate inference in such
networks, when an input is provided: different models and training
frameworks give rise to different ways of updating the parameters of the
model, but many of them require as a subroutine to perform iterative
inference to find a configuration of the hidden layers which is compatible
with the observed input.

\subsection{Energy-based models}

A particularly common type of non-temporal recurrent networks are those
whose dynamics correspond to minimizing an energy function or iteratively sampling
from the transition operator of a corresponding MCMC.
The energy function is associated with a joint distribution between visible units $\vv$
and latent units $\vh$, and we denote $\vs=(\vv,\vh)$ for
the joint random variable characterizing the whole state of the network.
For example, the energy function $E(\vs)$ can be associated with a joint distribution
via a Boltzmann distribution:
\begin{equation}
  P(\vs) = \frac{e^{-E(\vs)}}{Z}
\end{equation}
where $Z$ is the normalization constant or partition function associated with this
energy function. The energy function is parametrized as a sum involving coupling terms
$J_{i,j}(s_i,s_j)$ for the pair of units $i$ and $j$ taking the values $s_i$ and $s_j$.
There are also unary terms $U_i(s_i)$ which allow to bias the marginal distribution
of each unit, so the overall energy is
\begin{equation}
  E(\vs) = \sum_i U_i(s_i) + \sum_{i,j} J_{i,j}(s_i,s_j).
\end{equation}

In the discrete Boltzmann distribution, $s_i$ are binary-valued, $U_i = -s_i b_i$,
and $J_{i,j} = - W_{i,j} s_i s_j$, where $W$ is a symmetric matrix of weights.
Boltzmann machines with Gaussian (real-valued) units have a squared containing term for
each of the Gaussian units, such as $U_i = -\frac{s_i^2}{\sigma_i^2} - s_i b_i$.
The squared continuous states guarantee that the normalization constant exists (the integral does not diverge).
In continuous Hopfield networks, we also have $J_{i,j}=- W_{i,j} \rho(s_i) \rho(s_j)$ while
the unary terms involve an integral of the neuron non-linearity~\citep{Hopfield84}:
$U_i = \int_0^{\rho(s_i)} \rho^{-1}(s) ds$ (excluding biases for simplicity), where $s_i$ can be seen as the
voltage of neuron $i$ and $\rho$ as an element-wise neural transfer function
such as the sigmoid. Another interesting energy function is the one proposed
by~\citet{Bengio-arxiv2015} and \citet{Scellier+Bengio-arxiv2016}:
\begin{equation}
  \label{eq:s+b-energy}
  E(\vs) = \frac{1}{2}||\vs||^2 - \frac{1}{2}\sum_{i\neq j} W_{i,j} \rho(s_i) \rho(s_j) - \sum_i b_i \rho(s_i).
\end{equation}
These papers study this energy function in order to emulate back-propagation
in the recurrent network through a form of contrastive Hebbian learning which
corresponds to the standard spike-timing dependent plasticity (STDP).
However, they require some form of iterative inference
to approximately find a fixed point of
the neural dynamics that locally minimizes the energy function.
Furthermore, experiments reported by \citet{Scellier+Bengio-arxiv2016} suggest that the time needed for sufficient
convergence to the fixed point grows badly with the number of layers. Avoiding
such lengthy convergence of iterative inference motivates the work presented here.

\subsection{Iterative Inference}

In this paper we are most interested in the inference process by which low-energy
configurations of $\vh$ are obtained when the visible units $\vv$ are clamped
to some observed value $\vx$. In both Boltzmann machines and Hopfield networks,
inference proceeds by gradual changes to the state towards lower energy
configurations (and possibly some randomness injected). The most brain-like
inference is defined by a differential equation that specifies the temporal
evolution of neurons $s_i$:
\begin{equation}
  \label{eq:dsdt}
  \tau \frac{ds_i}{dt} = - \frac{\partial E(\vs)}{\partial s_i} + {\rm noise}
\end{equation}
where $\tau$ is a time constant, or $\epsilon=1/\tau$ is a learning rate for doing
(possibly stochastic) gradient descent in the energy. In Hopfield networks we
typically do not consider any injected noise (i.e., equivalently temperature is 0), but if noise is injected the resulting
stochastic process corresponds to Langevin dynamics (in discrete time, a Langevin
MCMC in a model associated with the given energy function at some temperature which
depends on the ratio of $1/\tau$ to the variance of the injected noise).
In the case of the continuous Hopfield network (and no noise), the differential equation corresponds
to a leaky neuron which integrates its inputs:
\begin{equation}
  \tau \frac{ds_i}{dt} = \rho'(s_i) \left( \sum_j W_{i,j} \rho(s_j) - s_i \right)\,.
\end{equation}
A fixed point is reached when
\begin{equation}
   s_i = \sum_j W_{i,j} \rho(s_j).
\end{equation}

When the architecture has no lateral connection ($W_{i,j}=0$ for units on the same layer),
it may be possible to speed up the inference process: instead of performing
gradient descent on all the units simultaneously, one can directly solve
for the value of $s_i$ for all units on the same layer that minimize the
energy, given the values of the units in the other layers. In particular,
if the only connections are between successive layers, it means that it is
possible to alternatively update all the odd layers and then all the even
layers, each time jumping to the minimum of the energy, conditioned on
the fixed layers. An example of this approach is developed by~\citet{Scellier+Bengio-arxiv2016}
for the energy in Eq.~\ref{eq:s+b-energy} and ``hard sigmoid''
or bounded rectification non-linearity $\rho(s)=\max(0,\min(1,s))$.
This provably~\citep{Scellier+Bengio-arxiv2016} gives rise to the following updates:
\begin{equation}
  \label{eq:direct}
   s_i \leftarrow \rho(\sum_{j\neq i} W_{j,i} \rho(s_j)). 
\end{equation}

In the case of the deep Boltzmann machine (which also has a layered architecture
with no lateral connections), the commonly used block Gibbs update is of the form
\begin{equation}
  \label{eq:gibbs-update}
  \vh_k \sim P(\vh_k | \vh_{k-1}, \vh_{k+1})
\end{equation}
where $P$ is derived from the energy function, and the above
updates the units $\vh_k$ in layer $k$ using the current values $\vh_{k-1}$ and $\vh_{k+1}$ of the
units in the layer below and above respectively (denoting $\vv=\vh_0$).

In general, once implemented in discrete time, iterative inference will be of the form
\begin{equation}
  \label{eq:general-update}
  \vh_k \leftarrow (1-\frac{1}{\tau}) \vh_k + \frac{1}{\tau} F_k(f_k(\vh_{k-1}),g_{k+1}(\vh_{k+1}),{\rm noise})
\end{equation}
where $f_k(\vh_{k-1})$ represents the bottom-up contribution into $\vh_k$, from the layer below, denoted $\vh_{k-1}$,
$g_{k+1}(\vh_{k+1})$ represents the top-down contribution into $\vh_k$, from the layer above, denoted $\vh_{k+1}$.
The notation $F_k$ indicates how the bottom-up, top-down and noise are combined. Typically $F_k(a,b,c)=a+b+c$
or the sum is followed by a non-linearity, or the output $F_k(a,b,c)$ is the result of sampling from a distribution
(e.g., for the Boltzmann machine with discrete units).
We can think of Eq.~\ref{eq:general-update} as implementing a transition operator for
a Markov chain whose stationary distribution is associated with some wanted energy function.
When $\tau=1$ the above is a direct update (like in the block Gibbs update of Eq.~\ref{eq:gibbs-update}
or jumping to the analytic solution of Eq.~\ref{eq:direct}), whereas when $\tau>1$ the layer $\vh_k$
gradually moves towards $F_k(f_k(\vh_{k-1}),g_{k+1}(\vh_{k+1}),0)$ (or wanders around it if noise is injected).
A direct update is very efficient and is guaranteed to down the energy when there are no lateral connections and only connections
between successive layers.
Stochastic versions use the injected noise argument to actually sample $\vh_k$ from a conditional
distribution. The $\vh_k$ argument is used to account for inertia in the update, e.g., a discrete-time
implementation of Eq.~\ref{eq:dsdt} would only gradually modify the value of $\vh_k$ towards the
value that the layers below and above want to see.
The differential equation form of Eq.~\ref{eq:general-update} is
\begin{equation}
  \label{eq:dhdt}
  \tau \frac{d\vh_k}{dt} = F_k(f_k(\vh_{k-1}),g_{k+1}(\vh_{k+1}),{\rm noise}) - \vh_k
\end{equation}
which has a 0-temperature fixed point at
\begin{equation}
 \vh_k = F_k(f_k(\vh_{k-1}),g_{k+1}(\vh_{k+1}),0).
\end{equation}
The feedforward contribution $f_k(\vh_{k-1})$ to layer $k$ is typically of the form
\begin{equation}
   f_{k,i}(\vh_{k-1}) = b_i + \sum_{j \in {\rm layer}\; k-1} W_{i,j} \rho(h_{k-1,j}).
\end{equation}
for unit $i$ of layer $k$, with bias $b_i$ and incoming weights $W_{i,j}$ with $j$ from layer $k-1$.
Similarly, the feedback contribution $g_{k+1}(\vh_{k+1})$ from layer $k+1$ into layer $k$ would
be of the form
\begin{equation}
   g_{k,i}(\vh_{k-1}) = b_i + \sum_{j \in {\rm layer}\; k+1} W_{i,j} \rho(h_{k+1,j}).
\end{equation}

\section{Sufficient Conditions for Fast Feedforward Approximate Inference}

Let us consider a layered architecture such as those discussed above,
in which the connections into each neuron are split into {\em groups},
which biologically may correspond to dendritic branches. For example,
we talk below about the group of bottom-up connections (from the lower layer)
and the group of top-down connections (from the upper layer).
Signals coming into the neuron from each group $b$ below may be modulated by
a different gain (the $\alpha_b$ below), so that
the recurrent dynamics are governed by the following differential
equation, for neuron $i$:
\begin{equation}
  \label{eq:multi-branch-update}
  \tau \frac{ds_i}{dt} = \sum_b \alpha_b (d_{b,i}(\vs) - s_i) + {\rm noise}
\end{equation}
where $s_i$ represents the average somatic voltage of neuron $i$ (averaging
over the short-term variations due to spikes),
the sum is over different dendritic branches into that neuron,
$d_{b,i}$ represents the {\em prediction} made by branch $b$ about the
state of neuron $i$, and $\alpha_b$ is the gain associated
with branch $b$ that arises from dendritic conductances
and $\tau = \sum_b \alpha_b$, following~\citet{Urbanczik+Senn-2014}. In the presence of synaptic bombardment that defines the so-called high-conductance state, as it is observed {\em in vivo} in some cortical areas, $\tau$ is short, on the order of a few milliseconds ~\citep{Destexhe2003}. When interpreting $d_{b,i}$ as dendritic voltages and
$s_i$ as a somatic voltage, their differences in (\ref{eq:multi-branch-update}) arises from the diffusion process these voltages are subject to in a branching cable~\citep{Koch2004}. The gains $\alpha_b$ appear as a dendritic coupling strength that, for conductance-based synaptic inputs and strongly bi-directional somato-dendritic coupling, become dynamic quantities that are mainly determined by the total excitatory and inhibitory synaptic conductance on branch $b$~\citep{Sacramento2016}. This could provide an additional mechanism
enabling a dynamical re-weighting of the contribution $d_{b,i}(\vs)$ of each dendritic branch to the somatic potential $s_i$. But such gains could also be
adapted on a slow timescale via branch strength plasticity~\citep{Losonczy2008}.
For simplicity, we take the dendritic branch gains $\alpha_b$ to be constant, which is a good approximation when the antidromic current flow from
the soma back to the dendritic compartments is small~\citep{Urbanczik+Senn-2014}.

We model $d_{b,i}(\vs)$ as the usual affine prediction
\begin{equation}
  \label{eq:affine-branch}
   d_{b,i} = c_{b,i} + \sum_j W_{b,i,j} \rho(s_j)
\end{equation}
where $W_{b,i,j}$ represents the synaptic weight from some other neuron $j$ into dendritic
branch $b$ of neuron $i$, and $c_{b,i}$ plays the role of branch-specific offset.
Below we consider the special case where there
are only two branches, a bottom-up branch ($b=0$) computing the feedforward activations
$f_k(\vh_{k-1})$ into layer $k$, from layer $k-1$, and a top-down branch computing the
feedback activations $g_{k+1}(\vh_{k+1})$ into layer $k$, from layer $k+1$. There
could also be dendritic branches accounting for lateral connections or from other areas. 
From a biological perspective, we can think of the dendritic branch receiving bottom-up connections
as the basal dendritic branch, while the dendritic branch receiving top-down connections
as the apical dendritic branch.
The basal and apical dendritic branches are well-studied in the case of cortical layer 5
pyramidal neurons, a very large and ubiquitous type of neuron \cite{Larkum2013}.

The 0-temperature fixed point of Eq.~\ref{eq:multi-branch-update} is obtained
by setting the left-hand side to 0 and solving:
\begin{equation}
  \label{eq:fp}
  s^*_i = \frac{\sum_b \alpha_b d_{b,i}(\vs^*)}{\sum_b \alpha_b}
\end{equation}
where $\vs^*$ is the fixed-point solution, a convex weighted sum of the contributions
coming from all the branches.
Let us compare this solution to the result of performing {\em only feedforward
  computation}, i.e.,
\begin{equation}
  \label{eq:ff-result}
  \vh_k = \vd_{0,k}(\vh_{k-1}) = f_k(\vh_{k-1})
\end{equation}
where $\vh_0=\vv$, and $\vd_{b,k}(\vh_{k-1})$ is the vector
containing the output of the feedforward (bottom-up) branch $d_{b,i}$ for all units $i$ in layer $k$,
and depending only on the activations of units in layer $k-1$.
To match Eq.~\ref{eq:ff-result} and Eq.~\ref{eq:fp} it is enough
to have
\begin{equation}
  \label{eq:ff-layerwise}
   d_{b,i}(\vs^*) = d_{0,i}(\vs^*)
\end{equation}
for all $b>0$, i.e., {\em all the dendritic branches agree on the
  value that the neuron should take}. Equivalently, the condition is that
\begin{equation}
  \label{eq:condition}
     d_{b,i}(\vs^*) = s^*_i.
\end{equation}
We call the above condition (in any form) the {\bf good mutual prediction condition}
because it means that each dendritic branch is outputting a value which agrees with
the values produced by the other branches.

In the case where there are two dendritic branches, one for bottom-up, feedforward connections
and one for top-down, feedback connections, this condition corresponds to having {\em consecutive layers
  forming a good auto-encoder}, as we show below. Let $\vd_{1,k}(\vh_{k+1}) = g_{k+1}(\vh_{k+1})$ represent the
contribution of the feedback connections from layer $k+1$ into layer $k$.
Then Eq.~\ref{eq:ff-layerwise} means that bottom-up contributions $f_k(\vh_{k-1})$ agree
with top-down contributions $g_{k+1}(\vh_{k+1})$:
\begin{equation}
  \vh_k = f_k(\vh_{k-1}) = g_{k+1}(\vh_{k+1})
\end{equation}
and thus
\begin{equation}
  \vh_k = f_k(\vh_{k-1}) = g_{k+1}(f_{k+1}(\vh_k))
\end{equation}
i.e., the feedforward and feedback connections of consecutive layers form a good auto-encoder:
in the case of feedforward and feedback dendrites, the good mutual prediction condition
is equivalent to a {\bf good auto-encoder condition}.

The consequence of the above analysis is that if the good mutual prediction condition
(or the good auto-encoder condition) is satified, then initializing the network
by the result of a pure feedforward computation sets it very close to the
fixed point of the 0-temperature network dynamics when the inputs are clamped to the observed value.
In the stochastic case (non-zero temperature), the feedforward initialization would
initialize the inference near a mode of the conditional distribution $P(\vh | \vv)$
associated with the inference task, which is very convenient.

\section{Synaptic Learning Rules Giving Rise to the Good Mutual Prediction Condition}

Now, why would consecutive layers form a good one-layer auto-encoder?
That clearly depends on the particulars of the training framework, but
several elements of existing learning algorithms for such networks conspire
to make successive layers good auto-encoders. For example, in the case
of restricted Boltzmann machines~\citep{Hinton06} trained with CD-1 (contrastive divergence with 1 step),
the weight update is 0 if the feedback weights perfectly reconstruct the input.

A related idea was discussed by Geoff Hinton in a recent talk~\citep{Hinton-Stanford-talk-27-04-2016}
(minute 44 of the video) in the context of a biologically plausible implementation of back-propagation
in multi-layer networks.
By having each pair of consecutive layers form a good auto-encoder, the feedback weights would not
perturb the activations computed in the feedforward pass, except to the extent that they would carry the perturbations
on the upper layer due to their having changed (from their feedforward value)
in the direction opposite to the error gradient. The idea of using the feedforward pass
to obtain a good initialization for fast inference in a deep Boltzmann machine pre-trained as a stack
of RBMs (where again each pair of consecutive layers typically forms a good autoencoder)
as also discussed by~\citet{Salakhutdinov+Hinton-2009-small}.

See the recirculation algorithm~\citep{Hinton+McClelland-NIPS1987}
and backprop-free auto-encoders by difference target-propagation~\citep{Lee+Bengio-NIPSDL2014-small,Lee-et-al-MLKDB2015-small}
for related ways to train consecutive pairs of layers so that they form a good auto-encoder,
without requiring explicit back-propagation into the encoder through the decoder.

The analysis in this paper suggests that in order to obtain the desirable
fast inference, the training framework should guarantee, either automatically
as a side effect of its objective, or via an additional term in the training objective,
that consecutive layers should form a good auto-encoder, or more generally that
different dendritic branches are trained to predict each other (or equivalently,
to predict the somatic voltage). This could be achieved with a local
objective function (for each dendritic branch $b$ of neuron $i$) of the form
\begin{equation}
     C_{b,i} = (s_i - d_{b,i}(\vs))^2
\end{equation}
that is minimized when Eq.~\ref{eq:condition} is satisfied. If $d_{b,i}$
follows the usual affine form of Eq.~\ref{eq:affine-branch}, then
the gradient of $C_{b,i}$ with respect to $W_{b,i,j}$ contains a 
term corresponding to updates of the form
\begin{equation}
  \label{eq:delta-W}
   \Delta W_{b,i,j} \propto (s_i - d_{b,i}(\vs)) \rho(s_j)
\end{equation}
where $s_i$ (the somatic voltage) acts like a target of a linear
regression with predictor $d_{b,i}(\vs)$, with inputs $\rho(s_j)$. This corresponds to a biological form of the classical error-correcting
rule~\citep{Widrow62}. However, Eq.~\ref{eq:delta-W}
does not take into account the possible indirect effect of $W_{b,i,j}$ on $C_{b,i}$
via the offect of $W_{b,i,j}$ on the fixed point of $s_j$, so future work should
investigate that.

The plasticity rule derived in Eq.~\ref{eq:delta-W} can also be seen in the context of dendritic predictions of somatic spiking~\citep{Urbanczik+Senn-2014}. In this biological version of the rule, the difference between the somatic and dendritic voltage is replaced by the difference between the instantaneous somatic spike rate and the rate predicted by the dendritic voltage. Both quantities can be read out by the dendritic synapse, the back-propagating action potential (bAP) and the local dendritic voltage.  Due to the near-Poisson
spiking of in vivo cortical neurons~\citep{Shadlen1998}, the instantaneous rate of action potentials
provides an unbiased estimate of the underlying somatic voltage $s_i$. There is in fact experimental evidence that plasticity depends on the postsynaptic voltage~\citep{Artola1990,Sjostrom2001} and bAPs~\citep{Markram1997}, see~\citep{Clopath2010} for a phenomenological model. Classical spike-timing dependent plasticity (STDP, \citet{Markram1997}) is reproduced by a spiking version of rule (\ref{eq:delta-W}) in a 1-compartment neuron~\citep{Brea2013}.


\begin{figure}[htpb]
  \begin{center}
    \hspace*{-1cm}\includegraphics[width=0.55\textwidth]{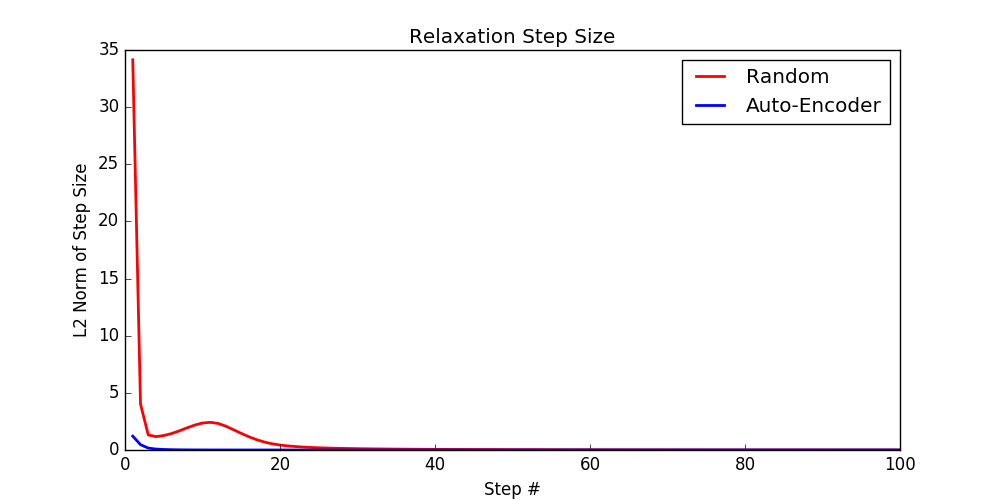} \hspace*{-1cm}
    \includegraphics[width=0.55\textwidth]{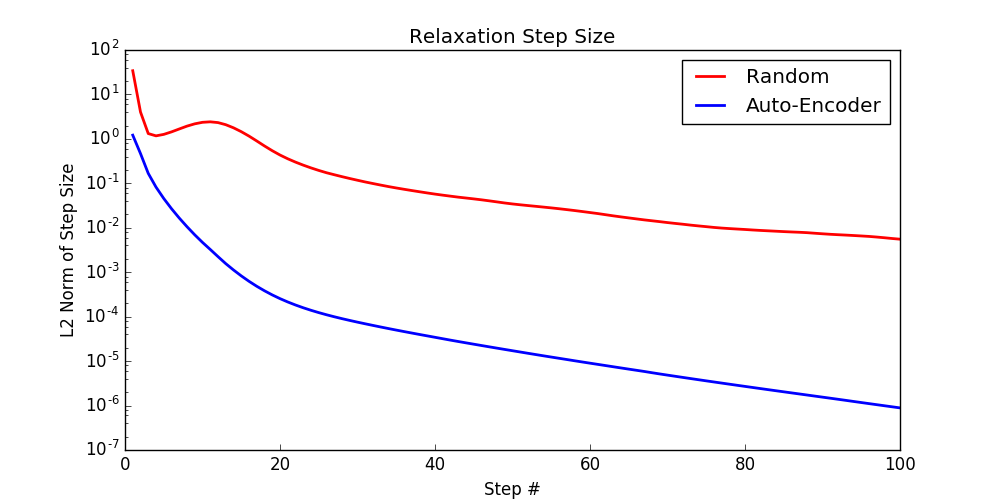}
  \end{center}
    \hspace*{-3cm}\begin{minipage}{1.4\textwidth}
  \caption{Convergence of the inference relaxation, with randomly set weights vs weights obtained
    by minimizing layerwise reconstruction error (auto-encoder). {\bf Left:} magnitude of the update steps in the
    space of the vector of all hidden layers state, after each update, vs number of updates. {\bf Right:} same in log-scale.}
  \label{fig:convergence-1}
    \end{minipage}
\end{figure}

\section{Simulation Results}

We have measured the convergence of relaxation to a fixed point in a recurrent network obeying a direct
version of Eq.~\ref{eq:multi-branch-update}, where odd layers or even layers are updated at each time step,
according to Eq.~\ref{eq:fp} (seen as a fixed point equation). It means that the bottom-up and top-down
contributions are averaged, except for the top hidden layer, which only has a bottom-up input.
The experiments are performed on the MNIST dataset and they compare different settings of the
weights.

Fig. \ref{fig:convergence-1} (left) shows that convergence is almost instanteneous when the consecutive layers form a
good auto-encoder, while the right of the figure shows that not only does it start closer to the fixed point but
it approaches it exponentially at a faster rate. That experiment compares randomly initialized weights
where the feedback weights equal the transpose of the feedforward weights with weights obtained by training
a stack of ordinary auto-encoders (with the piecewise-linear non-linearity $\rho(s)=\max(0,\min(1,s))$).
The neural network has 784 inputs and 3 hidden layers and we tried different hidden layer sizes 500 and 1000, with the
same results obtained.

\section{Conclusion}

We have proposed conditions under which a recurrent stochastic network would perform fast approximate inference
that is equivalent to running only a feedforward pass from inputs into deep hidden layers and shortcuts
the biological relaxation process. These conditions
would avoid the need for a lengthy iterative inference to either reach a fixed point or a stationary distribution
associated with the conditional distribution of hidden layers given a visible layer. This could be useful
both to speed-up training and using of such models, as well as a biologically plausible way to achieve
fast inference that matches well with recent successes obtained with feedforward neural networks trained
with back-propagation. The main ingredient of these assumptions is that different dendritic branches
predict the average of their mutual prediction, or in the case where there are only bottom-up
and top-down branches, each pair of successive layers
forms a good auto-encoder. Because a fixed point of this recurrent bottom-up top-down circuitry
can be explicitly calculate (approximately to the extent that the reconstruction error is small),
a single effective feedforward pass already lands very close to the fixed point and from this
the fixed point is reached very quickly.
Interestingly, the kind of training objective that enables this fast convergence property
matches recent proposals for synaptic update rules
in multi-compartment models of pyramidal cells~\citep{Urbanczik+Senn-2014}, with neural computation
differential equations that essentially correspond to the multi-branch dynamics
of Eq.~\ref{eq:multi-branch-update} and the updates of Eq.~\ref{eq:delta-W}.

However, a complete machine learning story is still missing. Keep in mind that there are many
sets of weights that can give rise to small reconstruction error.  This paper may shed light
on how inference could be performed efficiently, but more work is needed to build a biologically plausible
theory of learning that, from a machine learning perspective, explains how all the layers can be
trained together towards better fitting the observed data or rewards.
Recent work~\citep{Scellier+Bengio-arxiv2016} has shown how
inference in the kind of recurrent network discussed here (with computation
corresponding to minimizing an energy function) could be considerably
sped-up by only considering a small perturbation away from the positive
phase fixed point. However, the proposed mechanism still required to reach
a fixed point of the dynamics with inputs clamped. The assumptions introduced here
are sufficient conditions to make sure that this relaxation
could also be performed very fast. However, even augmented with fast inference, the
framework of ~\citet{Scellier+Bengio-arxiv2016}, although it suggests a way to ``propagate''
prediction errors into internal layers and do gradient descent on the weights, it
still only considers the deterministic scenario, still requires symmetric weights,
still only deals with the supervised case (point-wise predictions), and still does
not incorporate the sequential aspect of observed data.

Hence, although this contribution helps to deal with the issue of fast
approximate inference, a biologically plausible implementation of efficient supervised
learning (as in back-propagation) and unsupervised learning (as in Boltzmann machines) remains
a challenge for future investigations aiming to bridge the gap between deep learning
and neuroscience.

\iffinal
\section*{Acknowledgments}

The authors would like to thank Tong Che, Vincent Dumoulin, Kumar Krishna Agarwal
for feedback and discussions, as well as NSERC, CIFAR, Samsung and
Canada Research Chairs for funding.
\fi



\bibliographystyle{natbib}
\bibliography{strings,ml,aigaion,biblio,neurosci}

\end{document}